\documentclass{article}

\PassOptionsToPackage{numbers, sort&compress}{natbib}
\usepackage[final,dblblindworkshop]{neurips_2026}
\workshoptitle{Continual Learning in the Era of Foundation Models and Embodied Agents}

\usepackage{hyperref}
\hypersetup{
  pdftitle={Continual Adaptation Across Modalities: Learning and Forgetting in Unified Multimodal Models},
  pdfauthor={Xinrui Wang, Zhihao Chen, Jia Tang, Alexandra Gomez-Villa, Bart{\l}omiej Twardowski, Joost van de Weijer, Songcan Chen},
}
\usepackage{graphicx}
\usepackage{xcolor}
\usepackage{booktabs}
\usepackage{subcaption}
\usepackage{amsmath}
\usepackage{amssymb}

\usepackage{multirow}
\usepackage{array}
\usepackage{float}

\title{Continual Adaptation Across Modalities: Learning and Forgetting in Unified Multimodal Models}
\author{%
  {\bfseries\small Xinrui Wang$^{1,2}$, Zhihao Chen$^{3}$, Jia Tang$^{4}$, Alexandra Gomez-Villa$^{5}$,} \\
  {\bfseries\small Bart{\l}omiej Twardowski$^{6,2}$, Joost van de Weijer$^{5,2}$, Songcan Chen$^{1}$} \\[1mm]
  {\small $^{1}$Nanjing University of Aeronautics and Astronautics \quad $^{2}$Computer Vision Center, Barcelona} \\
  {\small $^{3}$EPFL \quad $^{4}$Shanghai Artificial Intelligence Laboratory \quad $^{5}$Universitat Aut\`onoma de Barcelona} \\
  {\small $^{6}$IDEAS Research Institute} \\
  {\small \texttt{wangxinrui@nuaa.edu.cn}} \\
}

\definecolor{brick}{RGB}{192,57,43}
\definecolor{darkgray}{RGB}{59,58,54}
\definecolor{clay}{RGB}{158,110,78} % top strip of fig:coupling_gradient
\definecolor{olive}{RGB}{137,148,122} % middle strip of fig:coupling_gradient
\definecolor{slateblue}{RGB}{111,127,140} % bottom strip of fig:coupling_gradient

\newcommand{\best}[1]{\textbf{#1}}

\graphicspath{{./plot/}}

\begin{document}

\maketitle

\begin{abstract}

Unified multimodal models (UMMs), which can process and generate diverse modalities within a single framework, have recently emerged as a powerful paradigm. However, when these models undergo continual learning on multiple tasks across different modalities, their learning and forgetting dynamics remain poorly understood. In this paper, we systematically investigate how architectural design and adaptation method shape intra-modal learning and cross-modal interference between the generation and understanding sides. Specifically, we report three findings. First, the severity of cross-modal interference decreases with the degree of modality-specific parameter isolation. Second, a within-model asymmetry: the generation side of a UMM is markedly more fragile than its understanding side. Third, what post-training adds is the first to be forgotten: purely post-training capabilities such as personalized generation collapse by 84\% to 90\%, while pretraining capabilities remain stable. Together, these results clarify how parameter isolation and post-training fragility shape UMMs' learning and forgetting dynamics, offering a starting point for developing continually evolving UMMs.

\end{abstract}

% ═══════════════════════════════════════════════════════════════════
\section{Introduction}
\label{sec:intro}
% ═══════════════════════════════════════════════════════════════════

Unified multimodal models (UMMs), which handle question answering, visual understanding, and image generation within a single framework, have recently emerged as a new paradigm~\cite{song2025bridge}. Representative systems include Chameleon~\cite{team2024chameleon}, Janus~\cite{wu2025janus, chen2025janus}, BAGEL~\cite{deng2025emerging}, Emu3~\cite{wang2024emu3}, and OmniDiffusion~\cite{li2026omni}. However, even these models cannot rely on pretraining alone: deployment demands that they continually learn new concepts and skills while retaining what they already know. This is far more complex for UMMs than for classic models, since a previous study~\cite{wei2026mitigating} showed that updating a UMM on a task in one modality can damage capabilities in another. Their study, however, was limited in two important respects. First, it considered only one architecture, a fully shared autoregressive transformer, whereas unified models today differ in parameter sharing (fully coupled to modality-specific expert routing) and generation (autoregressive or diffusion-based). Second, it treated generation as a static capability to be preserved, whereas in practice generation also evolves through continual learning, just as understanding does.

\begin{figure}[t]
  \centering
  \includegraphics[width=0.95\textwidth]{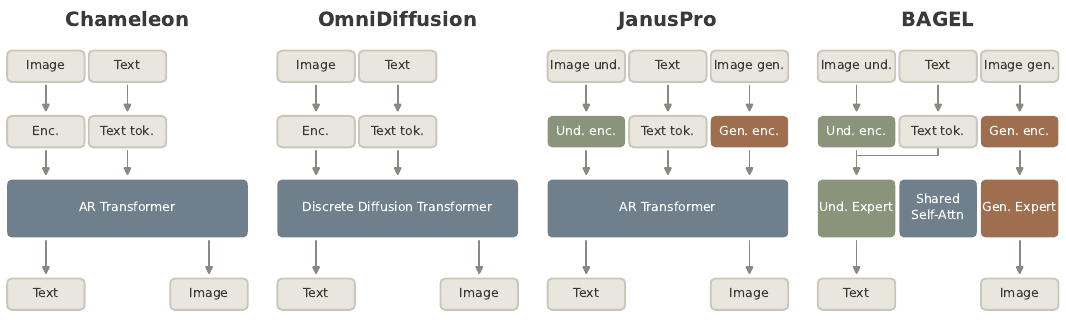}
  \caption{The four UMM architectures studied (Chameleon, OmniDiffusion, JanusPro and BAGEL), redrawn in simplified unified style and ordered by parameter coupling (high to low).}
  \label{fig:architectures}
\end{figure}

In this paper, we design a continual learning experiment spanning both understanding and generation tasks. For understanding, we fine-tune models sequentially on seven vision-language benchmarks covering multiple domains and capabilities~\cite{zhao2025mllm} (six VQA tasks and one image captioning task, listed in Section~\ref{sec:prelim}) and measure how well each model's performance on earlier tasks is retained. For generation, we focus on continual personalized image generation, where the model genuinely learns and generates new visual concepts. Four UMMs spanning the coupling spectrum and generation paradigms are studied (Figure~\ref{fig:architectures}); their architectures are detailed in Section~\ref{sec:prelim}. In the analysis, we track both intra-modal performance changes, as in conventional continual learning settings~\cite{chen2024coin,guo2025hide,liao2025continual,smith2024continual}, and cross-modal interference between personalization and understanding training. Our experiments reveal three findings, further developed in Section~\ref{sec:inter}, Section~\ref{sec:intra} and Appendix.

\textbf{Finding 1: Parameter isolation reduces cross-modal interference.} The severity of cross-modal interference decreases with the degree of modality-specific parameter isolation. As shown in Figure~\ref{fig:coupling_gradient}, DreamBooth~\cite{ruiz2023dreambooth} personalization degrades understanding most in the fully coupled Chameleon ($3.6$\,percentage points (pp) on average), less in OmniDiffusion ($2.4$\,pp), and almost not at all in JanusPro and BAGEL; understanding training likewise degrades Chameleon's personalized generations the most, with BAGEL at nearly zero and the other two models in between.

\textbf{Finding 2: UMMs' generation is more fragile than their understanding.} After the same sequential understanding tasks, abilities are still preserved (from $-4.7$~pp of forgetting for BAGEL to $-10.4$~pp for Chameleon), while personalized generation can be largely destroyed. After the sequential personalization tasks, nearly all models can no longer preserve the semantics of concepts learned early, with the mean Q-Align score dropping by 84\% to 90\% from its initial value.

\textbf{Finding 3: What post-training adds is the first to be forgotten.} Within the sequential understanding tasks, forgetting concentrates on exactly the tasks that training improved the most relative to zero-shot: for Chameleon, the three tasks with the largest gains (Medical, Remote Sensing, and Autonomous Driving, each improved by roughly 45~pp) also lose the most (11.5 to 18.3~pp), while tasks with smaller gains are nearly retained; JanusPro shows the same trend almost monotonically (from 0.1~pp of forgetting for a 1.6~pp gain to 17.3~pp for a 45.8~pp gain). The generation side shows the same asymmetry: personalized generation, a purely post-training capability, collapses while generic generation is barely affected by either understanding or personalization training (from 5.8~pp for Chameleon to near zero for JanusPro and BAGEL).

% ═══════════════════════════════════════════════════════════════════
\section{Preliminaries}
\label{sec:prelim}
% ═══════════════════════════════════════════════════════════════════

To cover the design space of current UMMs, we select four representative architectures spanning the modality coupling spectrum, a key axis that distinguishes existing UMMs (Figure~\ref{fig:architectures}): Chameleon~\cite{team2024chameleon} is fully coupled with autoregressive decoding; OmniDiffusion~\cite{li2026omni} shares an equally coupled backbone but generates images via masked discrete diffusion over MAGVIT-v2 tokens; JanusPro~\cite{chen2025janus} decouples the visual encoders (SigLIP for understanding, VQ tokenizer for generation) over a shared autoregressive transformer; BAGEL~\cite{deng2025emerging} decouples at the transformer level through a Mixture-of-Transformer-Experts, where all tokens share self-attention but understanding and generation activate separate FFN experts, with autoregressive text decoding and rectified-flow image generation. To study these models under cross-modal continual learning, we design a framework of tasks spanning both understanding and generation.

\noindent\textbf{Understanding.} We fine-tune sequentially on seven vision-language benchmarks spanning distinct domains and capabilities, six VQA tasks (Finance, Science, Medical, Remote Sensing, Autonomous Driving, and OCR) and one image captioning task (vizwiz), and measure retention of earlier tasks at the end of the sequence, with MMBench and MMMU as additional evaluation points to track how general understanding evolves. Understanding-side adaptation follows the classic supervised fine-tuning (SFT) recipe, with the task-specific loss adapted to each model's generation scheme.

\noindent\textbf{Generation.} We study continual personalized image generation on a sequence of 19 visual concepts, each represented by 6 to 10 images~\cite{an2026unictokens}, and track generation quality over time. For personalization we consider the two mainstream schemes: DreamBooth~\cite{ruiz2023dreambooth}, which adapts model parameters, and progressive prompt tuning (YoPrompt), a text-inversion-style scheme that optimizes a soft prompt instead. Because parameter-modifying updates affect the model far more, the main text focuses on DreamBooth, and the prompt-based variant is deferred to Appendix~\ref{sec:initial_gen_quality}. All parameter-modifying adaptation uses LoRA~\cite{hu2022lora} with per-model hyperparameters selected by grid search, so observed differences reflect architecture rather than optimization.

\noindent
In the experiments, we report accuracy for VQA tasks, SPICE score for captioning~\cite{anderson2016spice}, and CLIP-I, CLIP-T~\cite{radford2021learning}, Q-Align, and Q-Quality~\cite{zhang2025q} for personalized generation (described in Appendix~\ref{app:setup}). In the following, we first focus on cross-modal interference (Section~\ref{sec:inter}) and then on intra-modal learning and forgetting (Section~\ref{sec:intra}).

% ═══════════════════════════════════════════════════════════════════
\section{More Parameter Sharing Increases Cross-Modal Interference}
\label{sec:inter}
% ═══════════════════════════════════════════════════════════════════

\begin{figure}[t]
  \centering
  \begin{minipage}{0.49\textwidth}
    \centering
    \includegraphics[width=\textwidth]{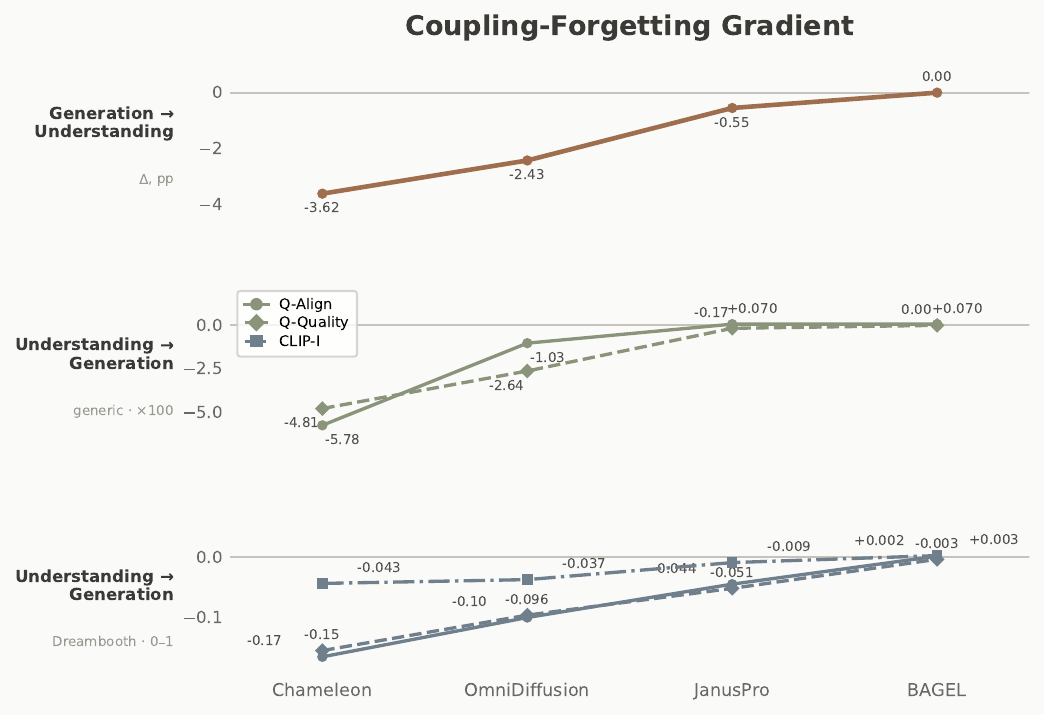}
  \end{minipage}\hfill
  \begin{minipage}{0.49\textwidth}
    \centering
    \includegraphics[width=\textwidth]{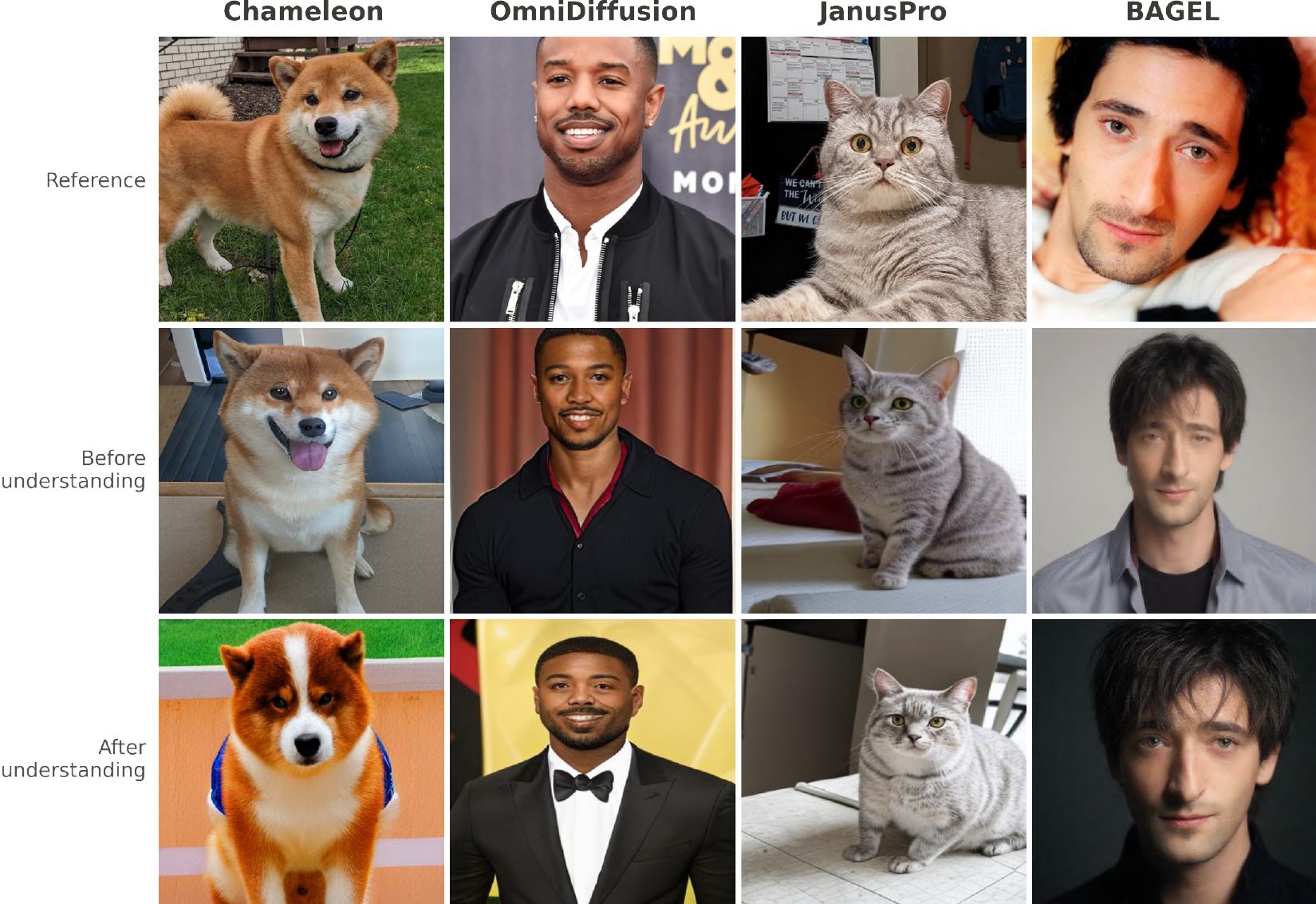}
  \end{minipage}
  \caption{Cross-modal interference in UMMs. \textbf{Left:} Inter-modal interference. \textcolor{clay}{Top}: generation to understanding. \textcolor{olive}{Middle}: understanding to generic generation. \textcolor{slateblue}{Bottom}: understanding to personalized generation.  \textbf{Right:} personalized concept generations before and after the seven sequential understanding tasks (rows: reference image, generation before and after understanding training). }
  \label{fig:coupling_gradient}
\end{figure}

In UMMs this interference is bidirectional: personalizing a model for generation changes its understanding, and understanding training changes its generation. 

\noindent\textbf{Generation to Understanding.} To measure this direction, we average each model's performance change over all understanding tasks trained so far, computed after its continual personalization training. As shown in Figure~\ref{fig:coupling_gradient} (\textcolor{clay}{top}), personalizing Chameleon degrades its understanding by $3.6$\,pp on average; OmniDiffusion drops $2.4$\,pp; JanusPro loses almost nothing; and BAGEL, whose Mixture-of-Transformer-Experts keep understanding and generation updates fully separate over shared attention, is not affected at all. The more modality-specific parameters are isolated, the less generation-side updates perturb the shared representations that understanding relies on. Masked discrete diffusion moderates the damage further, likely because its loss covers only masked image positions and is spread across timesteps, so individual updates exert less pressure on the shared backbone. Per-task records for each model are in Appendix~\ref{sec:und_after_db}.

\noindent
To rule out adapter placement and capacity as the driver of this gradient, we report where the DreamBooth update lands in each model. In Chameleon, OmniDiffusion, and JanusPro, the entire LoRA ($39.9$M, $43.1$M, and $37.4$M trainable parameters, respectively) sits in the shared backbone, with no LoRA on any adapter or alignment module; in BAGEL, the update is confined to the generation-specific attention projections ($12.6$M). The interference levels in Figure~\ref{fig:coupling_gradient} therefore track the architecture rather than the adapter placement or its capacity: identically placed, similarly sized updates degrade Chameleon's understanding by $3.6$\,pp but leave JanusPro's almost untouched, and BAGEL's smaller, fully generation-specific update leaves understanding unaffected by construction.

\noindent\textbf{Understanding to Generation.} In the reverse direction the same trend holds (Figure~\ref{fig:coupling_gradient}, \textcolor{olive}{middle} and \textcolor{slateblue}{bottom}): understanding training degrades Chameleon's personalized generations the most (Q-Align $\Delta -16.5$\,pp) and leaves BAGEL's decoupled per-concept generations at $\Delta \approx 0$, with the other two models in between. Full decoupling almost removes the cross-modal interference entirely, but it is a double-edged sword for a self-evolving UMM: while decoupling shields each side from the other's updates, it also severs the cross-modal synergy that motivates unified models in the first place. The full per-model record for personalized generation is in Appendix~\ref{sec:initial_gen_quality}. Moreover, generic generation is far less affected than personalized generation: understanding training degrades it by only Q-Align $\Delta -5.8$\,pp for Chameleon and near zero for JanusPro and BAGEL (detailed in Appendix~\ref{sec:gen_after_und}), whereas personalized generation loses Q-Align $\Delta -16.5$\,pp for Chameleon. The most fragile capabilities are precisely those acquired through post-training, while generic generation, largely a pretraining ability, remains stable.

% ═══════════════════════════════════════════════════════════════════
\section{Intra-Modal Learning and Forgetting}
\label{sec:intra}
% ═══════════════════════════════════════════════════════════════════

We now turn to intra-modal learning and forgetting: whether the seven sequentially trained understanding tasks are retained, and how generation quality changes along the personalization sequence.

\begin{figure}[t]
  \centering
  \includegraphics[width=0.9\textwidth]{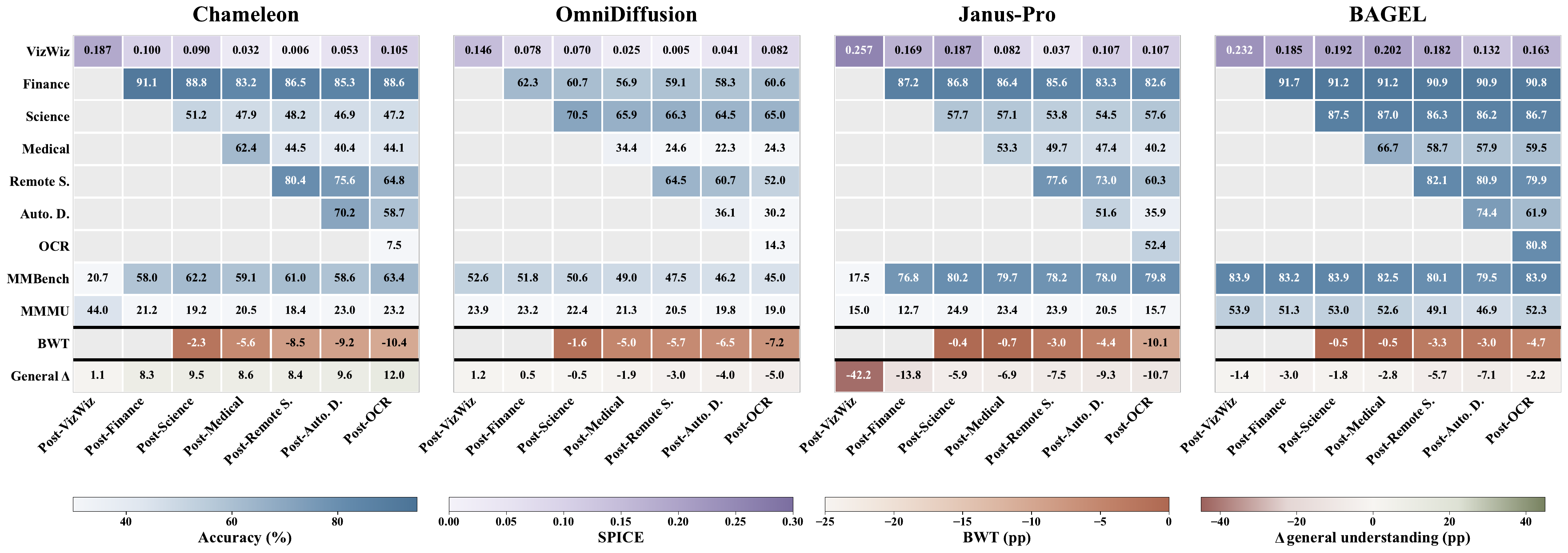}
  \caption{\textbf{Intra-modal performance matrices (Understanding).} Lower-triangle score matrices across the 7 continual understanding tasks: each column shows the state after one more trained task, so the main diagonal reflects forward plasticity and entries below it reflect retention (backward transfer). The MMBench/MMMU rows track general understanding via zero-shot probes, and the BWT row summarizes forgetting on the trained tasks, averaging the final score minus the score right after that task's training (the diagonal). The probe row reports each probe's final score minus its zero-shot baseline, tracking general understanding.}
  \label{fig:under_degrade}
\end{figure}

\begin{figure}[t]
  \centering
  \includegraphics[width=0.9\textwidth]{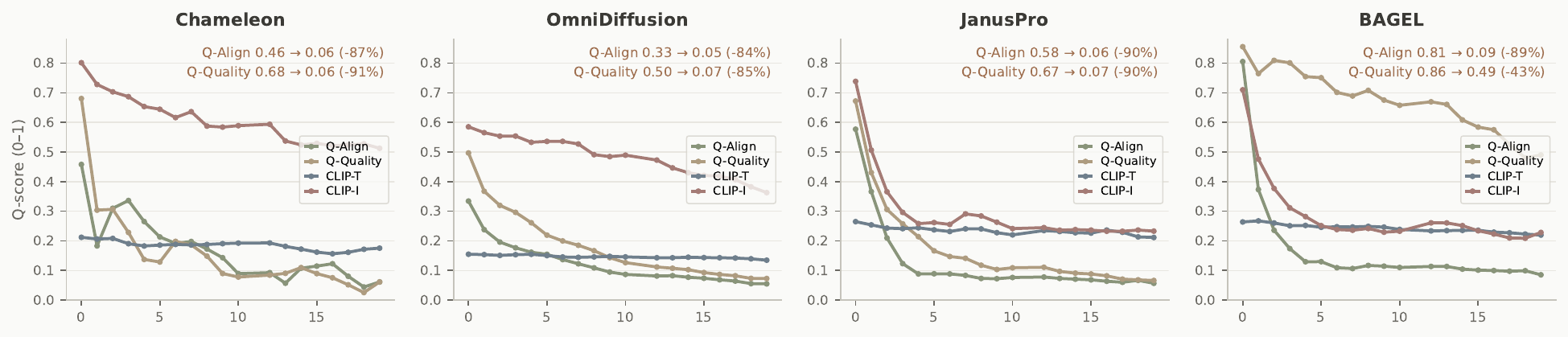}
  \caption{Mean generation quality of the first personalized concept (Q-Align, Q-Quality, CLIP-T, and CLIP-I) as the remaining 18 DreamBooth concepts are learned in sequence, per model.}
  \label{fig:fragility}
\end{figure}

\textbf{Sequential Understanding.} Figure~\ref{fig:under_degrade} summarizes retention across the seven-task sequence. BAGEL forgets least (BWT $-4.7$\,pp). OmniDiffusion degrades almost as much as Chameleon (BWT $-7.2$\,pp). Chameleon and JanusPro lose the most (BWT $-10.4$ and $-10.1$\,pp, respectively). Zero-shot MMBench and MMMU probes show that Chameleon, starting from weak zero-shot performance (Table~\ref{tab:understanding}), is first hurt by the captioning task (MMBench from 37.63 to 20.70) and then improves as the task sequence progresses, ending at 63.36, its peak. JanusPro's MMBench is first crushed by the captioning task (17.53 after vizwiz) and then recovers to within 0.6\% of its peak, while its MMMU declines from 24.92 to 15.72, consistent with fine-tuning on the captioning task harming instruction-following ability. Chameleon's OCR is also the clearest case of a task ending below its zero-shot level: its OCR diagonal (7.54) falls below the already weak zero-shot OCR (10.40, Table~\ref{tab:understanding}), as interference from the six earlier tasks---most of all the text-dense captioning task---leaves the OCR update with more to repair than it can recover. JanusPro's OCR improves over zero-shot (49.37 to 52.40), while BAGEL's dips only slightly (85.58 to 81.10). Per-model retention patterns are detailed in Appendix~\ref{sec:seq_und}.

\textbf{Sequential Generation.} The generation side is much more fragile than understanding. Figure~\ref{fig:fragility} and \ref{fig:seq_gen_change} (Appendix~\ref{sec:db_per_concept}) track the generation quality of the first personalized concept as 19 DreamBooth concepts are learned in order: quality decays in every model as the sequence progresses, and for most models the decline is drastic, with Q-Align falling from $0.46$ to $0.06$ for Chameleon and from $0.81$ to $0.09$ for BAGEL, and with drops of 84\% to 90\% across models (e.g.\ Q-Align by $87\%$ for Chameleon and $89\%$ for BAGEL). Q-Quality collapses similarly in the coupled models (from $0.68$ to $0.06$ for Chameleon, $0.50$ to $0.07$ for OmniDiffusion, and $0.67$ to $0.07$ for JanusPro) but far less in BAGEL ($0.86$ to $0.49$, $-43\%$), whose generation expert keeps producing visually plausible images even as the concept identity is overwritten. The near-total collapse is specific to these semantic-consistency scores: identity and text alignment erode far more slowly, with CLIP-I falling from $0.80$ to $0.51$ for Chameleon, $0.59$ to $0.36$ for OmniDiffusion, $0.74$ to $0.23$ for JanusPro, and $0.71$ to $0.23$ for BAGEL, and CLIP-T nearly flat throughout (e.g., $0.21$ to $0.18$ for Chameleon). Unlike understanding, generation can be destroyed outright by a short sequence of concepts: although there are more concepts than understanding tasks (19 vs.\ 7), each concept is trained on far less data. Each concept is compressed into dense, identity-specific weights from only a handful of reference images, so later updates overwrite earlier ones; understanding tasks, in contrast, are spread over large training sets and a shared feature space that supports retention.

% ═══════════════════════════════════════════════════════════════════
% \section{Conclusion}
% \label{sec:conclusion}
% % ═══════════════════════════════════════════════════════════════════

% We studied learning and forgetting in four UMMs spanning the coupling spectrum. Two findings emerge: a coupling-forgetting gradient, where interference decreases monotonically with modality-specific parameter isolation, and a within-model asymmetry, where generation degrades to near-zero while understanding merely declines. Self-evolving UMMs should isolate modality-specific parameters and treat generation as the binding constraint; diffusion-based training is left to future work.
\small
\bibliographystyle{plainnat}
\bibliography{references}

\clearpage 
% ═══════════════════════════════════════════════════════════════════
\appendix
% ═══════════════════════════════════════════════════════════════════

\section{Implementation Details}
\label{app:setup}

\noindent
All models are adapted with LoRA~\cite{hu2022lora}; learning rate, LoRA rank and $\alpha$ are the optimal configurations selected by grid search for each model, so that observed differences reflect architecture rather than optimization. q/k/v/o denote the self-attention projections, gate/up/down the feed-forward projections, and aligner the MLP adapter after JanusPro's visual encoders. Tables~\ref{tab:impl_und},~\ref{tab:impl_gen}, and~\ref{tab:impl_eval} summarize the training and evaluation settings. OmniDiffusion follows the same seven-task and 19-concept sequences with the same LoRA recipe; its results are reported as aggregated statistics (means and SEMs).

\noindent
\textbf{DreamBooth personalization.} Each concept is trained from its 6 to 10 reference images together with a concept identifier token and negative samples for training stability; the 19 concepts are learned sequentially into a single LoRA, so that later concepts overwrite earlier ones. We find that this sequential training needs additional samples unrelated to any concept to stay stable; without them, personalization tends to fail, and the model often loses the ability to learn new concepts entirely after two or three concepts. Generation is evaluated with simple templates such as ``a photo of \texttt{<concept>}'' and its variants.

\noindent
\textbf{Progressive prompt tuning (YoPrompt).} Following Yo'Chameleon~\cite{nguyen2025yo}, YoPrompt encodes each concept as a 16-token learnable soft prompt while keeping the model frozen. Because only 6 to 10 reference images are available per concept, negative samples are mined from a large-scale image corpus: images are ranked by their CLIP image similarity~\cite{radford2021learning} to the mean feature of the concept's reference images, and the most similar ones are kept as high-quality hard negatives to regularize training. The prompt length is then allocated progressively, following Yo'Chameleon's adaptive prompt length strategy: negatives with higher similarity receive more learnable tokens, while the full token budget is reserved for the true positive images, so the model can distinguish the target concept while still exploiting shared features. This negative-sample design is critical for Chameleon in particular: on the three concepts where high-quality negatives are available, its Q-Align and Q-Quality reach 0.520 and 0.485, well above the 19-concept averages of 0.386 and 0.370 obtained with the default selection. For Chameleon, JanusPro, and BAGEL the prompt is initialized from the concept's textual description; for OmniDiffusion, textual initialization proved insufficient, so the prompt is initialized from image-patch clusters obtained with UniCTokens~\cite{an2026unictokens}.

\begin{table}[H]
  \centering
  \caption{Sequential understanding: seven tasks trained in a fixed order, each on its training split. Within each model, all seven tasks share one configuration.}
  \label{tab:impl_und}
  \setlength{\tabcolsep}{4pt}
  \small
  \begin{tabular}{llllll}
    \toprule
    Model & Learning rate & LoRA rank & LoRA $\alpha$ & Epochs & LoRA target modules \\
    \midrule
    Chameleon     & 1e-4         & 16           & 32          & 1 & q/k/v/o \\
    OmniDiffusion & 2e-5     & 16         & 32        & 1 & q/k/v/o \\
    JanusPro 7B   & 2e-4     & 16         & 32        & 1 & q/k/v/o + aligner \\
    BAGEL         & 2e-5     & 32         & 64        & 1 & q/v/o/gate/up/down (no k\_proj) \\
    \bottomrule
  \end{tabular}
\end{table}

\begin{table}[H]
  \centering
  \caption{Personalized generation: continual concept-by-concept training on 6 to 10 images per concept~\cite{an2026unictokens}, with a fixed number of steps per concept. YoPrompt optimizes a 16-token soft prompt instead of LoRA (marked --).}
  \label{tab:impl_gen}
  \setlength{\tabcolsep}{4pt}
  \small
  \begin{tabular}{lllcccccc}
    \toprule
    Model & Method & Concepts & Token num & Learning rate & LoRA rank & LoRA $\alpha$ & Steps \\
    \midrule
    Chameleon     & Dreambooth & 19 & -- &  3e-5      & 16      & 32      & 100 \\
    Chameleon     & YoPrompt   & 19 & 16 & 1e-3      & --    & --    & 4000 \\
    OmniDiffusion & Dreambooth & 19 & -- & 2e-5  & 16    & 32    & 800 \\
    OmniDiffusion & YoPrompt   & 19 & 16 & 1e-3  & --    & --    & 6000 \\
    JanusPro 7B   & Dreambooth & 19 & -- & 1e-5  & 16    & 32    & 1000 \\
    JanusPro 7B   & YoPrompt   & 19 & 16 & 5e-3  & --    & --    & 4000 \\
    BAGEL         & Dreambooth & 19 & -- & 1e-5  & 16    & 32    & 1000 \\
    BAGEL         & YoPrompt   & 19 & 16 & 1e-3  & --    & --    & 500 \\
    \bottomrule
  \end{tabular}
\end{table}

\begin{table}[H]
  \centering
  \caption{Evaluation protocols.}
  \label{tab:impl_eval}
  \begin{tabular}{lll}
    \toprule
    Axis & Metric & Protocol \\
    \midrule
    Understanding, VQA     & Accuracy (\%)       & six VQA benchmarks \\
    Understanding, caption & SPICE $\times 100$ & vizwiz \\
    Understanding, general & Accuracy (\%)       & MMBench, MMMU as evaluation points \\
    Generation             & CLIP-I, CLIP-T     & identity and text alignment \\
    Generation             & Q-Align, Q-Quality & semantic consistency and visual quality \\
    \bottomrule
  \end{tabular}
\end{table}

\noindent
\textbf{Evaluation metrics.} SPICE~\cite{anderson2016spice} scores captions by comparing the scene graph of the generated caption against those of the reference captions, rewarding generated captions that preserve the semantic propositions (objects, attributes, and relations) of the references; it is used for the vizwiz captioning task. CLIP-I and CLIP-T~\cite{radford2021learning} measure the identity and text alignment of generated images in CLIP embedding space: CLIP-I is the cosine similarity between the generated image and the concept's reference images (how well the subject identity is preserved), and CLIP-T the similarity between the generated image and the prompt text (how faithfully the generation follows the prompt). Q-Align and Q-Quality~\cite{zhang2025q} are two scores of the same LLM-based scorer. The scorer assigns logits to five quality categories (Excellent, Good, Fair, Poor, Bad), which we combine into a $0$--$1$ score by softmax-weighted averaging with category weights $1.0$, $0.75$, $0.5$, $0.25$, and $0.0$: Q-Align rates the semantic consistency between the generated image and the text description, and Q-Quality rates the visual quality of the generated image.

\section{Detailed Results}
\label{app:results}

\subsection{Understanding}

\subsubsection{Initial Understanding Quality}
\label{sec:initial_und_quality}

\begin{table}[H]
  \centering
  \caption{Understanding across all seven tasks and the general benchmarks.}
  \label{tab:understanding}
  \resizebox{\textwidth}{!}{%
  \begin{tabular}{lcccccccccc}
    \toprule
    Model & vizwiz & Fin & Sci & Med & RS & AD & OCR & MMBench & MMMU & Avg (VQA) \\
    \midrule
    \textbf{Bagel}        & \best{16.00} & 60.07 & \best{86.32} & \best{46.84} & \best{49.42} & \best{29.04} & \best{85.58} & \best{87.98} & \best{52.60} & \best{59.55} \\
    JanusPro 7B           & 9.31 & \best{68.02} & 56.11 & 25.03 & 31.80 & 16.35 & 49.37 & 84.82 & 32.18 & 41.11 \\
    Chameleon             & 7.47 & 57.55 & 9.41 & 16.70 & 35.04 & 24.69 & 10.40 & 37.63 & 24.89 & 25.63 \\
    OmniDiffusion         & 3.76 & 48.06 & 40.83 & 26.22 & 35.45 & 24.85 & 5.07 & 50.94 & 23.11 & 30.08 \\
    \bottomrule
  \end{tabular}}
\end{table}

\noindent
Table~\ref{tab:understanding} reports the zero-shot understanding capability of each model: all scores are measured on the pretrained checkpoints before any task-specific training. Averaged over the six VQA tasks (vizwiz is excluded because SPICE has a different scale), BAGEL is the strongest performer (59.55\%), followed by JanusPro 7B (41.11\%), OmniDiffusion (30.08\%), and Chameleon (25.63\%). As general understanding benchmarks, MMBench and MMMU probe broad multimodal reasoning across many capabilities rather than any single task, and are used here as evaluation points rather than training tasks; BAGEL again leads (87.98 and 52.60) and is the only model above 50\% on MMMU, while JanusPro 7B is close on MMBench (84.82) but drops to 32.18 on MMMU. Within the six VQA tasks, JanusPro 7B tops Finance but is comparatively weak on Medical and Autonomous Driving, hinting at uneven domain coverage. OmniDiffusion occupies the middle ground yet collapses on the two text-dense tasks, OCR and vizwiz. Chameleon trails the most: despite strong pretrained generation, its understanding is the weakest (below 20 on Science, OCR, and Medical), and the post-training gains reported in the main text are correspondingly the largest.

\subsubsection{Sequential Understanding: Per-Model Patterns}
\label{sec:seq_und}

\noindent
Bagel shows near-perfect retention: its upper-triangle structure is preserved and lower-triangle pre-training evaluations confirm stable baselines, with MMBench and MMMU flat. JanusPro's retention is nearly as poor as Chameleon's: its BWT of $-10.1$~pp matches Chameleon's $-10.4$~pp, with early tasks substantially overwritten by later ones. Chameleon degrades significantly on the early tasks (vizwiz, OCR, and AD), whose capabilities are partially overwritten by later tasks.

\subsubsection{Understanding under Continual Generation Adaptation}
\label{sec:und_after_db}

\noindent
\textit{(Understanding $\to$ Personalization: models first trained on 7 understanding tasks, then personalized on 19 concepts.)}

\begin{figure}[H]
  \centering
  \includegraphics[width=\textwidth]{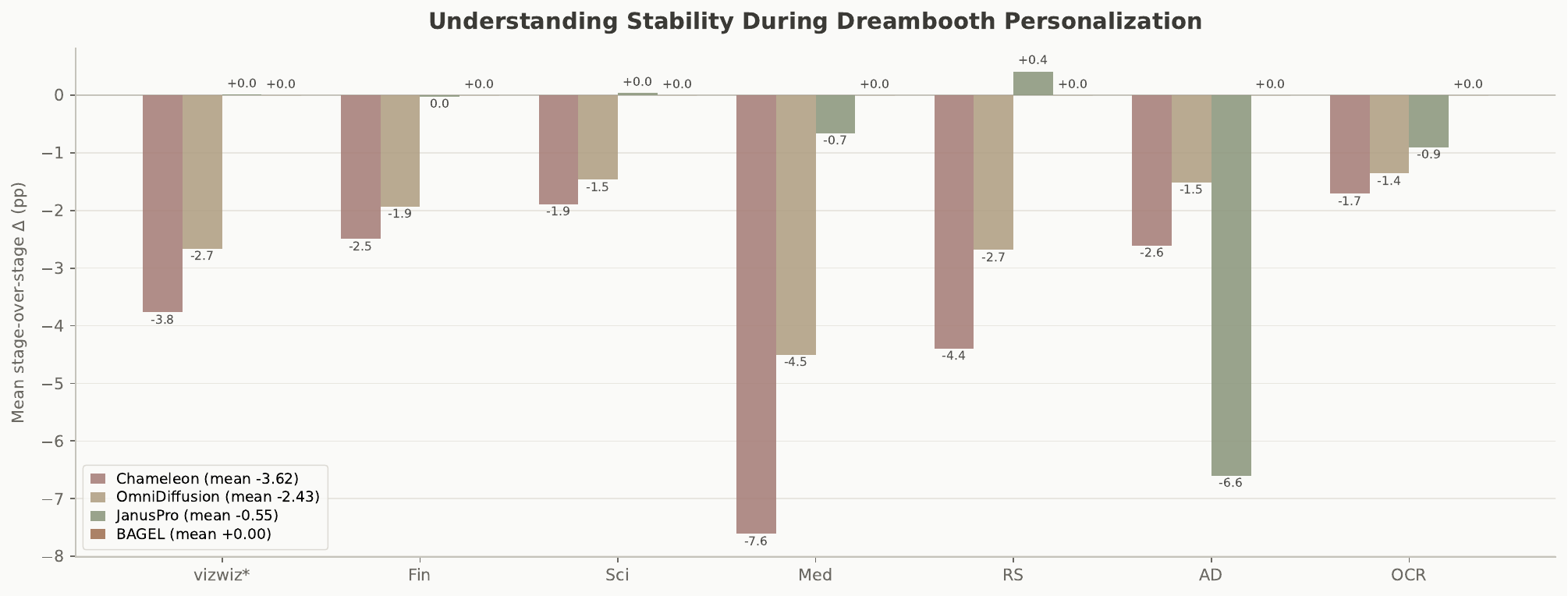}
  \caption{Per-task mean of the stage-over-stage $\Delta$ at and after each task's own training stage.}
  \label{fig:db_degrade_under_summary}
\end{figure}

\noindent
Since only DreamBooth-style personalization updates the model parameters, we focus on it here. Figure~\ref{fig:db_degrade_under_summary} plots the per-task mean deltas for Chameleon, JanusPro, and BAGEL.

\textbf{Bagel.} All deltas are exactly zero: personalization updates are isolated within the Generation Expert, and pure understanding never generates VAE tokens, so the Understanding Expert and shared attention produce identical outputs. The MoT architecture keeps generation-side updates fully isolated from the understanding side in practice.

\noindent
\textbf{JanusPro 7B.} Personalization mildly \textit{improves} most benchmarks on average (Fin $+0.9$, Med $+2.0$, RS $+1.4$~pp mean per stage), with AD as the exception ($-2.4$~pp), reflecting concept-benchmark interference.

\noindent
\textbf{OmniDiffusion.} Sharing a fully coupled backbone with Chameleon, its degradation is intermediate (mean $-2.4$~pp per stage, Figure~\ref{fig:coupling_gradient}), clearly weaker than Chameleon's $-3.6$~pp, consistent with masked discrete diffusion moderating the interference.

\noindent
\textbf{Chameleon.} Systematic negative deltas across most stages: understanding progressively degrades, and the zero-shot probes confirm the trend, with MMBench degraded by $-4.6$~pp on average and MMMU oscillating near zero.

\subsection{Generation}

\subsubsection{Generation under Continual Understanding Adaptation}
\label{sec:gen_after_und}

\begin{figure}[ht]
  \centering
  \includegraphics[width=\textwidth]{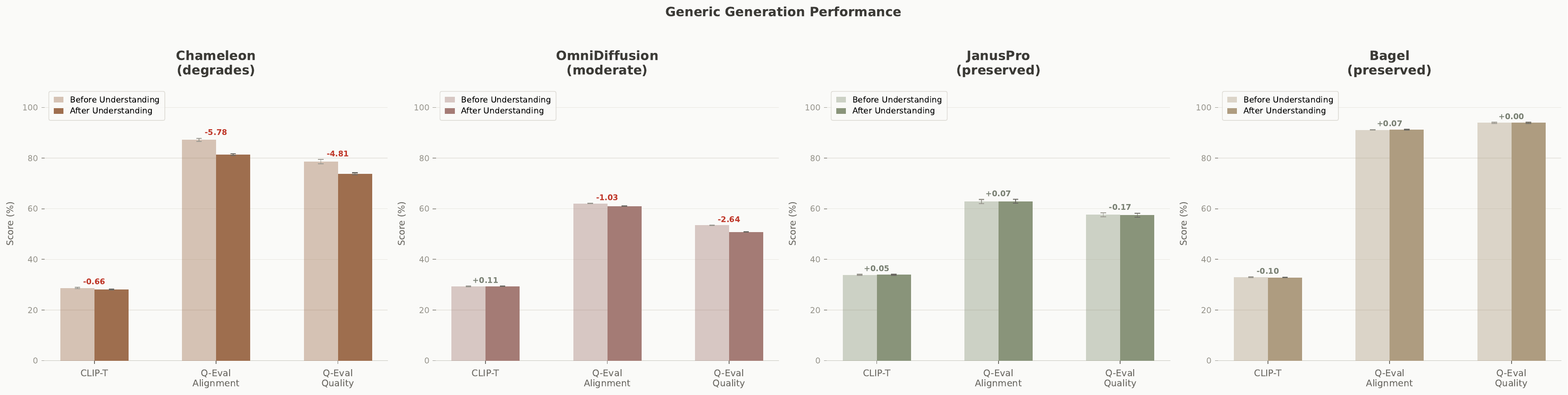}
  \caption{Generic generation performance before vs.\ after understanding, averaged across all concepts. Evaluated on 100 complex-conditioned text-to-image prompts (generated by an LLM).}
  \label{fig:under_degrade_generalgen}
\end{figure}

\begin{figure}[ht]
  \centering
  \includegraphics[width=\textwidth]{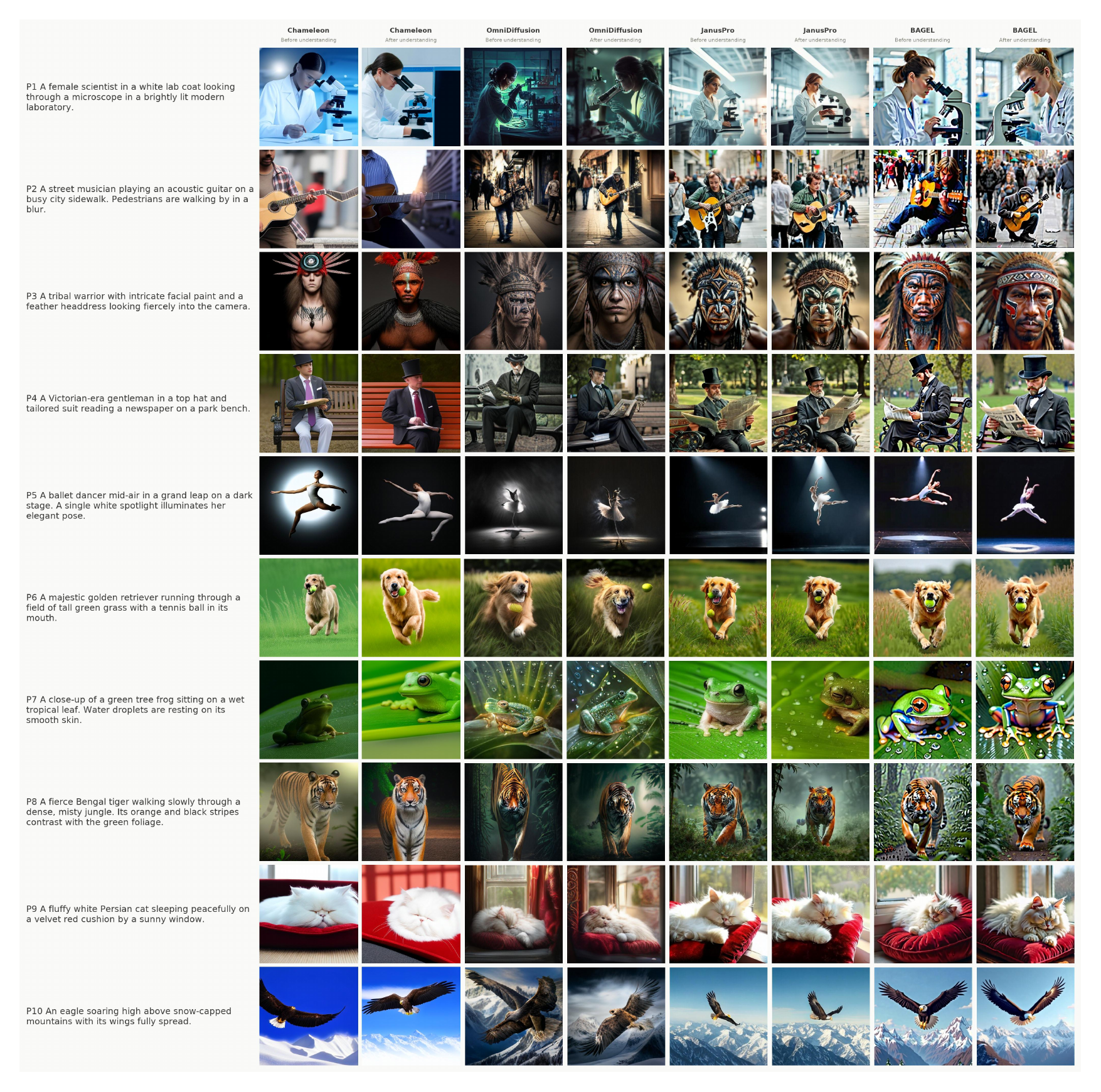}
  \caption{Generic generation before vs.\ after the seven understanding tasks. Each row shows one of the 10 representative prompts from the 100-prompt evaluation set, printed in full at the row head; columns are the four models, each with before- and after-understanding panels.}
  \label{fig:generic_after_under}
\end{figure}

\noindent
\textbf{Quantitative.} Figure~\ref{fig:under_degrade_generalgen} compares each model's generic generation before and after the seven understanding tasks on three metrics (CLIP-T, Q-Align, Q-Quality, all $\times 100$), with error bars showing the SEM. Chameleon degrades most: Q-Align drops from 87.20 to 81.42 ($\Delta -5.78$), Q-Quality from 78.62 to 73.81 ($\Delta -4.81$), while CLIP-T barely moves ($\Delta -0.66$), suggesting the degradation is primarily semantic and perceptual rather than a loss of textual alignment. OmniDiffusion, sharing the fully coupled backbone but trained with masked discrete diffusion, degrades less but still noticeably (Q-Align $\Delta -1.03$, Q-Quality $\Delta -2.64$, CLIP-T $\Delta +0.11$), consistent with its intermediate position in the interference trend of Figure~\ref{fig:coupling_gradient}. JanusPro and BAGEL are virtually unaffected: every metric stays within $\pm 0.2$ of its before-understanding value, and BAGEL's Q-Quality does not move at all.

\noindent
\textbf{Qualitative.} Figure~\ref{fig:generic_after_under} shows the same comparison for 10 representative prompts of the 100-prompt evaluation set, one row per prompt with the full prompt text at the row head. Chameleon's after-understanding panels show visible drift: fine-grained attributes in the prompt (objects, lighting, and spatial relations) are simplified or dropped, matching the Q-Quality drop above. JanusPro and BAGEL are indistinguishable from their before-understanding panels, and OmniDiffusion shows only minor deviations in a few prompts. The qualitative ranking thus mirrors the quantitative deltas and the architecture-level coupling order.

\subsubsection{Personalized Generation under Continual Understanding Adaptation}
\label{sec:initial_gen_quality}

\begin{figure}[H]
  \centering
  \begin{subfigure}{0.49\textwidth}
    \centering
    \includegraphics[width=\textwidth]{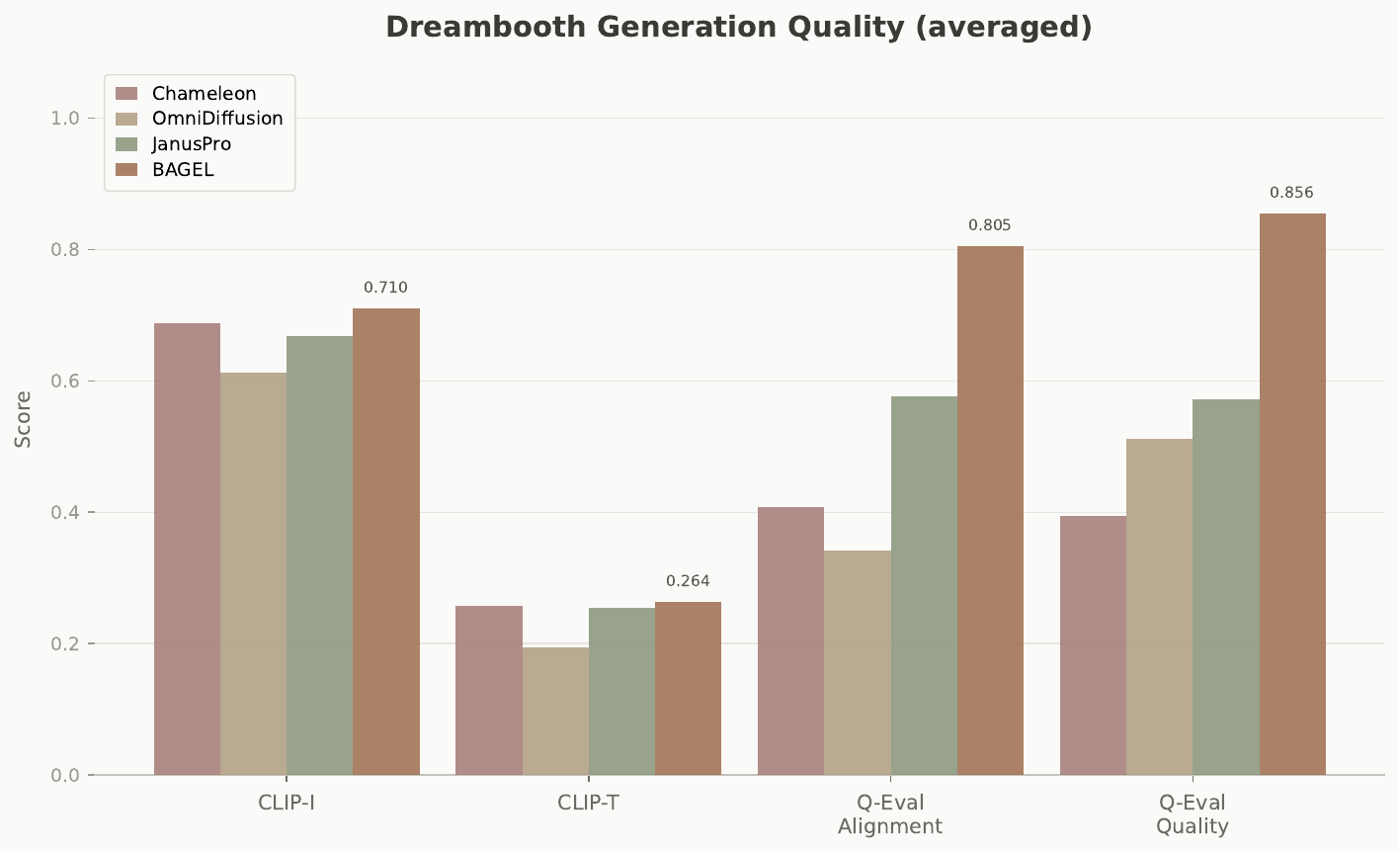}
    \caption{DreamBooth.}
    \label{fig:db_quality}
  \end{subfigure}\hfill
  \begin{subfigure}{0.49\textwidth}
    \centering
    \includegraphics[width=\textwidth]{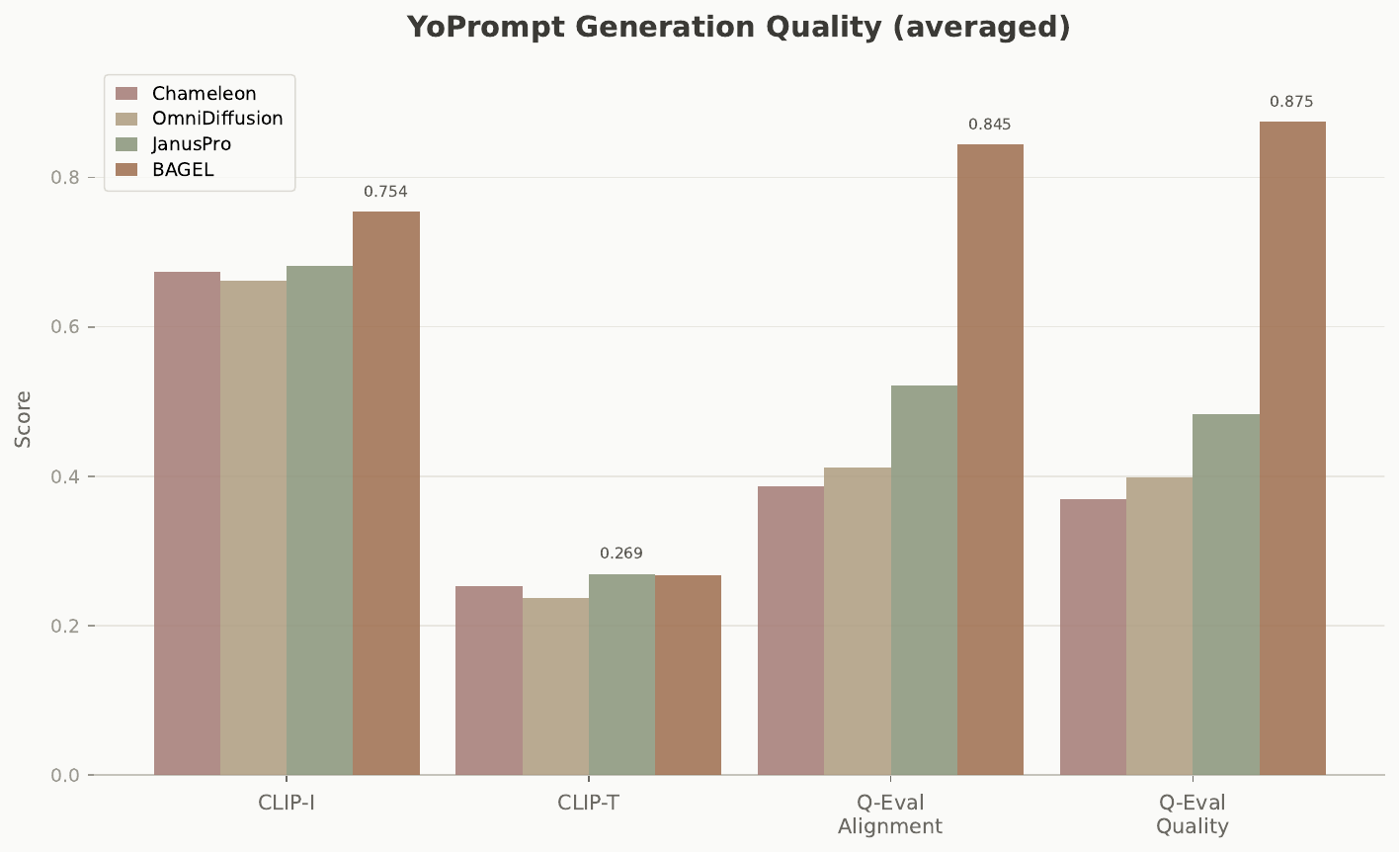}
    \caption{YoPrompt.}
    \label{fig:yoprompt_quality}
  \end{subfigure}
  \caption{Personalized generation quality, averaged across all 19 concepts.}
  \label{fig:gen_quality}
\end{figure}

\begin{figure}[H]
  \centering
  \includegraphics[width=\textwidth]{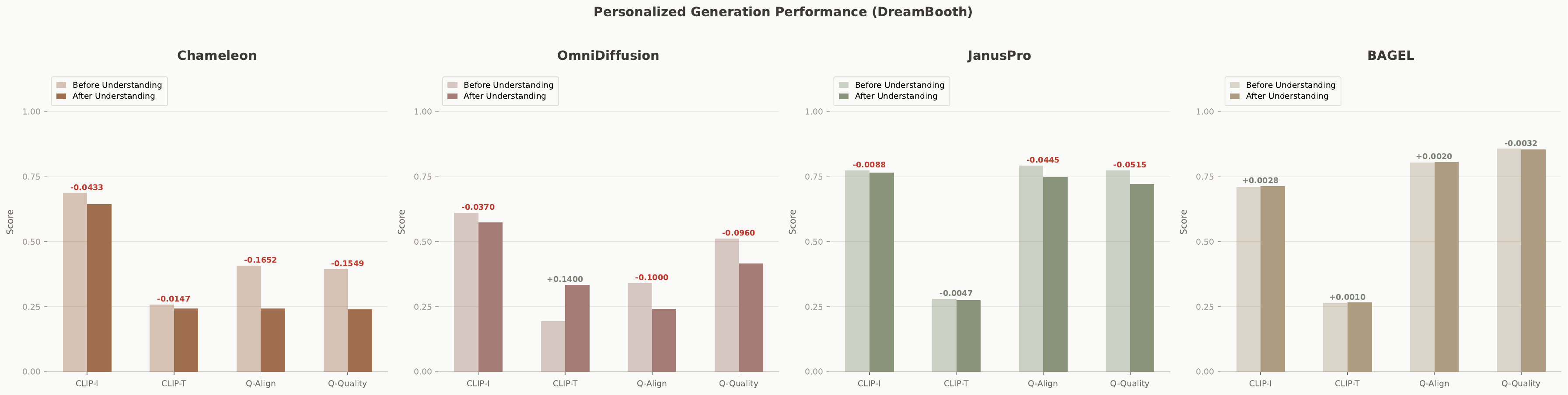}
  \caption{DreamBooth personalized generation before vs.\ after the seven understanding tasks, per model and metric (0--1 scale). Chameleon and JanusPro average 3 concepts (pre to T6, no re-personalization); BAGEL and OmniDiffusion average 19 concepts.}
  \label{fig:db_after_under}
\end{figure}

\begin{figure}[H]
  \centering
  \includegraphics[width=\textwidth]{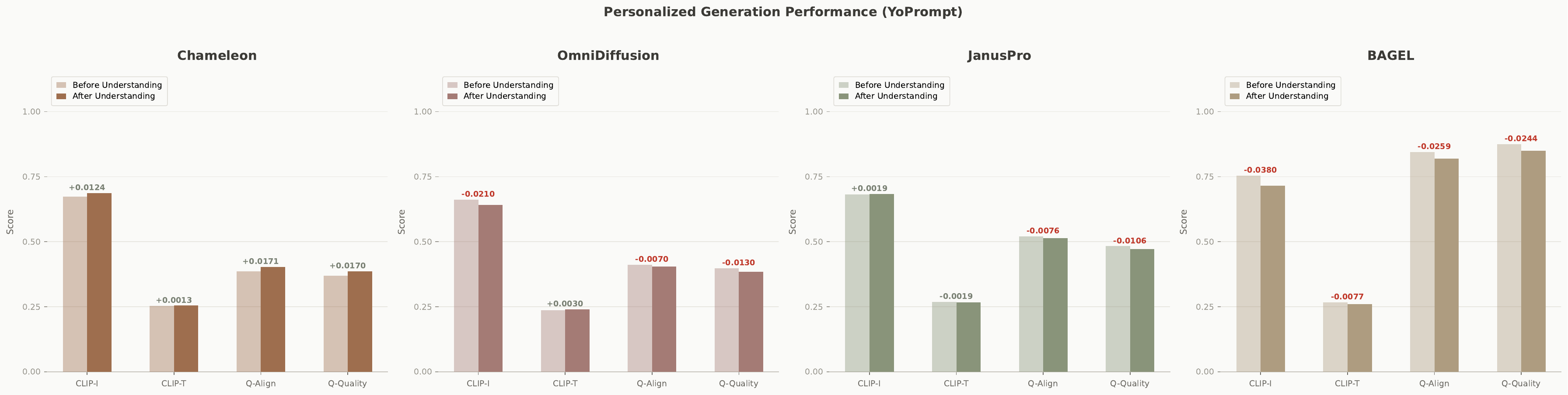}
  \caption{YoPrompt personalized generation before vs.\ after the seven understanding tasks, per model and metric (0--1 scale), averaged over 19 concepts.}
  \label{fig:yoprompt_after_under}
\end{figure}

\noindent
\textbf{Generic generation change after understanding.} Identity (CLIP-I~/~CLIP-T) is comparable across models in both schemes (CLIP-T slightly lower for OmniDiffusion), and the quality ranking is consistent: BAGEL $\gg$ JanusPro $\gg$ Chameleon $\approx$ OmniDiffusion on Q-Align and Q-Quality, with BAGEL's Q-Quality at 0.856 (DreamBooth) and 0.875 (YoPrompt) versus 0.37 to 0.40 for Chameleon and OmniDiffusion. Comparing the two schemes, prompt tuning is competitive with parameter updates: YoPrompt improves on DreamBooth for BAGEL (Q-Align 0.805 $\to$ 0.845) and especially for OmniDiffusion (0.341 $\to$ 0.412), while DreamBooth stays ahead for JanusPro (0.577 vs.\ 0.521) and Chameleon (0.409 vs.\ 0.386). Note that these measurements are per-concept \textit{independent} generations: each concept is personalized and evaluated separately, without the sequential training of Section~\ref{sec:intra}, whose cross-task degradation is reported in the main text. Chameleon's YoPrompt quality depends on negative-sample quality, analyzed in Appendix~\ref{app:setup}.

\noindent
\textbf{Personalization after understanding.} Figures~\ref{fig:db_after_under} and~\ref{fig:yoprompt_after_under} track these qualities through the seven understanding tasks.

\noindent
\textbf{DreamBooth.} Degradation follows the coupling order of Figure~\ref{fig:coupling_gradient}: Chameleon's Q-Align falls from 0.41 to 0.24 ($\Delta -0.165$) and Q-Quality from 0.39 to 0.24 ($\Delta -0.155$), with CLIP-I and CLIP-T dropping by $0.043$ and $0.015$; OmniDiffusion loses $0.100$ and $0.096$ on Q-Align and Q-Quality (its CLIP-T rises by $0.14$); JanusPro drops $0.045$ and $0.051$; and BAGEL stays within $0.003$ of zero on every metric. Personalization written into the model parameters is overwritten by later understanding training, in proportion to how much of the backbone it shares.

\begin{figure}[ht]
  \centering
  \includegraphics[width=0.9\textwidth]{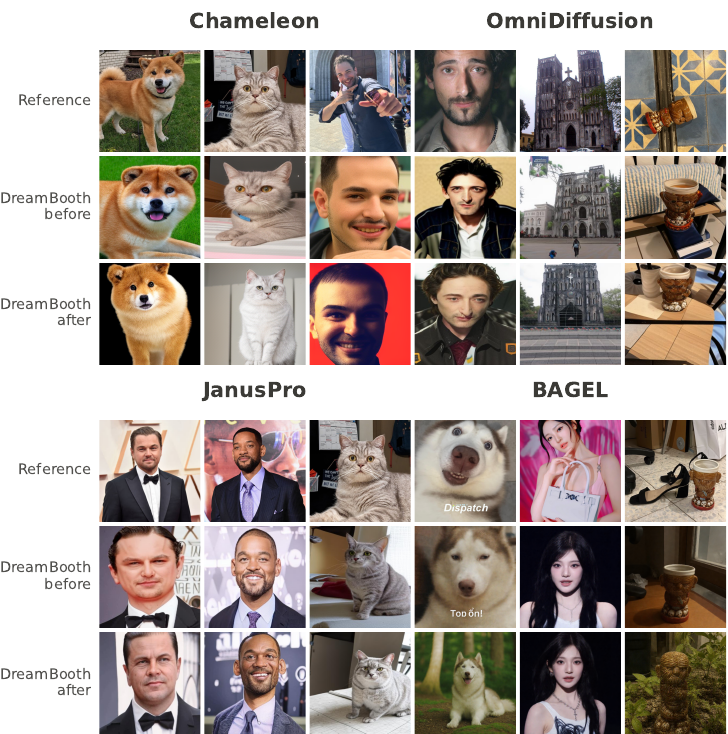}
  \caption{Per-concept DreamBooth generations before vs.\ after the seven understanding tasks, all four models (three concepts each, coupling order), reference image on top of each column.}
  \label{fig:pers_db}
\end{figure}

\noindent
\textbf{YoPrompt.} YoPrompt adds no parameter-level knowledge and otherwise relies on the model's base generation capabilities, so it behaves like generic generation and is essentially unaffected for Chameleon, OmniDiffusion and JanusPro: Chameleon's mean deltas are slightly positive on all four metrics ($+0.001$ to $+0.017$), and the other two stay within $0.021$ and $0.011$ of zero respectively. BAGEL is the exception: all four deltas are slightly negative, with CLIP-I at $\Delta -0.038$ (the largest), which is somewhat puzzling given that its DreamBooth deltas are negligible. We can offer only a tentative explanation: YoPrompt depends more heavily on the understanding side than parameter-based personalization, because the soft prompt is an input token sequence that, under BAGEL's hard routing, is processed by the updated understanding expert, while the generation expert that decodes it stays frozen. The effect is small, at most a few pp, and our data do not allow a definitive attribution.

\begin{figure}[ht]
  \centering
  \includegraphics[width=0.9\textwidth]{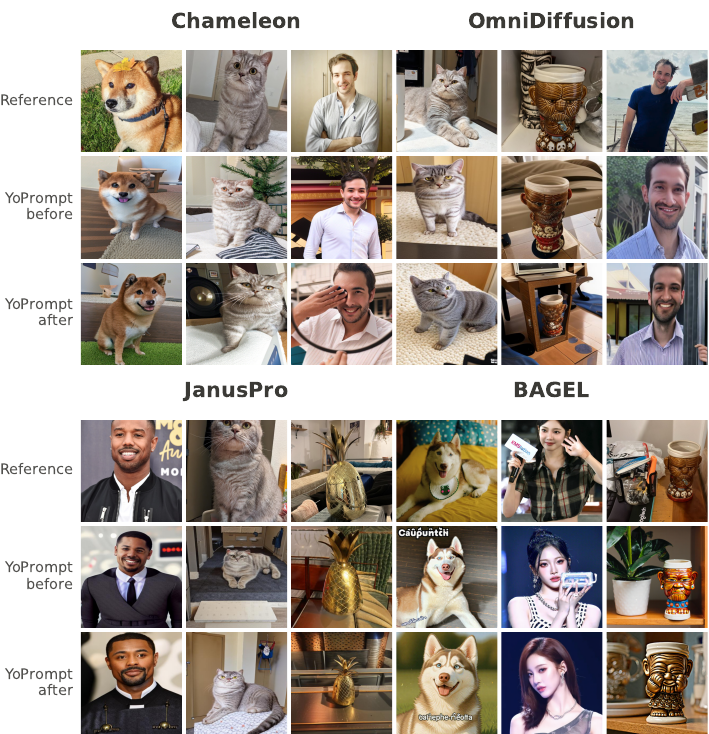}
  \caption{Per-concept YoPrompt generations before vs.\ after the seven understanding tasks, all four models (three concepts each, coupling order), reference image on top of each column.}
  \label{fig:pers_yoprompt}
\end{figure}

\noindent
This method dependence sharpens Finding~3: what post-training adds is first forgotten when it is written into parameters, whereas prompt-level personalization, which builds on the base model, remains largely stable apart from the small BAGEL exception discussed above.
\noindent
\textbf{Per-concept views.} Figures~\ref{fig:pers_db} and~\ref{fig:pers_yoprompt} show per-concept generations before and after the seven understanding tasks for all four models, with the reference image on top of each column; the columns are grouped per model in coupling order from Chameleon to BAGEL.

\subsubsection{Sequential Generation}
\label{sec:db_per_concept}

\begin{figure}[ht]
  \centering
  \includegraphics[width=0.99\textwidth]{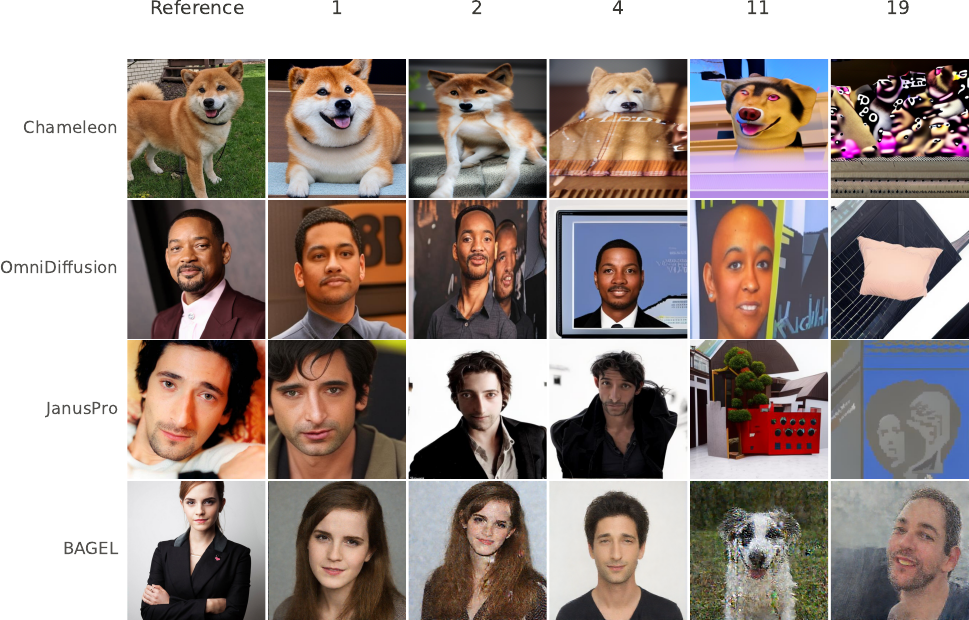}
  \caption{\textbf{Sequential generation, visualized.} Generated images of the first personalized concept at successive DreamBooth checkpoints: after the 1st concept (the target itself, bo for Chameleon, brody for JanusPro, emma for BAGEL and will for OmniDiffusion) and after the 2nd, 4th, 11th and 19th concepts of the sequence were learned. Rows are models, with the reference image on the far left of each row.}
  \label{fig:seq_gen_change}
\end{figure}

\noindent
Bagel's personalized generation is architecturally immune to the understanding-side updates of Section~\ref{sec:gen_after_und}, yet even BAGEL is destroyed by sequential DreamBooth training. Figure~\ref{fig:fragility} tracks the mean generation quality of the first personalized concept as further concepts are learned, for all four models: Q-Align falls from $0.46$ to $0.06$ for Chameleon ($87\%$), from $0.81$ to $0.09$ for BAGEL ($89\%$), from $0.58$ to $0.06$ for JanusPro ($90\%$), and from $0.33$ to $0.05$ for OmniDiffusion ($84\%$), a drop of 84\% to 90\% in every model. The MoT architecture confines each update to the Generation Expert, which is what protects BAGEL from understanding-side updates, but the first concept's identity lives in that same expert and is overwritten by later concept updates; the same holds for the coupled backbones of the other three models. Since YoPrompt optimizes only a prompt without updating the model's main parameters, it does not suffer from sequential-training interference, so this section reports DreamBooth only. Figure~\ref{fig:seq_gen_change} visualizes the first personalized concept's generations at successive checkpoints for each model.

\pagebreak
% ═══════════════════════════════════════════════════════════════════
\begin{ack}
This work was supported by the National Science and Technology Major Project of China (No. 2024YFB3311401), the National Natural Science Foundation of China (NSFC) under Grant No. 62376126 and No. 62576168, and the Open Project Funds for the Joint Laboratory of Spatial Intelligent Perception and Large Model Application (SIPLMA-2024-YB-05).

This research has also received funding from the European Union's Horizon Europe research and innovation programme under grant agreement No. 101214398 (ELLIOT). Furthermore, this work was supported by the Spanish Ministry of Science and Innovation (MCIN/AEI/10.13039/501100011033) and the European Regional Development Fund (FEDER) under project PID2022-143257NB-I00, and the grant RYC2021-032765-I.

Funding was also provided by the National Science Centre (NCN, Poland) under Grant No. 2023/51/D/ST6/02846, and the Funding for Outstanding Doctoral Dissertation in NUAA (BCXJ25-21).
\end{ack}

\end{document}